\pdfoutput=1

\documentclass[11pt]{article}

\usepackage{EMNLP2023}

\usepackage{times}
\usepackage{latexsym}

\usepackage[T1]{fontenc}

\usepackage[utf8]{inputenc}

\usepackage{microtype}

\usepackage{inconsolata}

\usepackage{nopageno}
\usepackage{epsf,epsfig,graphics}
\usepackage{amssymb,amsmath}
\usepackage{hyperref}
\usepackage{mathtools}
\usepackage{graphicx}
\usepackage{amsfonts}
\usepackage{booktabs}
\usepackage{siunitx}
\usepackage{array, multirow}
\usepackage[toc,page]{appendix}
\usepackage{graphicx}

\title{An Expectation-Realization Model for Metaphor Detection}

\author{Oseremen O. Uduehi \\
    School of EECS \\
    Ohio University \\
    Athens, OH 45701 \\
  \texttt{ou380517@ohio.edu} \\\And
  Razvan C. Bunescu \\
  Department of Computer Science \\
  University of North Carolina at Charlotte \\
  Charlotte, NC 28223 \\
  \texttt{razvan.bunescu@uncc.edu} \\}

\begin{document}
\maketitle
\begin{abstract}
We propose a metaphor detection architecture that is structured around two main modules: an expectation component that estimates representations of literal word expectations given a context, and a realization component that computes representations of actual word meanings in context. The overall architecture is trained to learn expectation-realization (ER) patterns that characterize metaphorical uses of words. When evaluated on three metaphor datasets for within distribution, out of distribution, and novel metaphor generalization, the proposed method is shown to obtain results that are competitive or better than state-of-the art. Further increases in metaphor detection accuracy are obtained through ensembling of ER models.
\end{abstract}

\section{Introduction and Motivation}

Metaphors are pervasive in everyday communication, as well as in creative writing such as novels and poetry. Metaphors enhance the communicative aspects of language by connecting concepts from new domains, often abstract, with more familiar ones, usually concrete \cite{LakoffJohnson80}. Metaphorical expressions have many uses, from helping frame an issue in order to emphasize some aspects of reality \cite{Boeynaems:IJoC17}, to creating a strong emotional effect \cite{blanchette_analogy_2001,citron_metaphorical_2014}. The ubiquity of metaphors means their computational treatment \cite{veale_metaphor_2016} has received significant attention in the NLP community, as surveyed by \citet{shutova_design_2015} and more recently \citet{tong-etal-2021-recent}. A distinction is made in the literature between {\it conventional} metaphors, which are entrenched in the conceptual system, and {\it novel} metaphors, which are unfamiliar. Owing to its important communicative function, metaphorical expression detection has been approached over the years using a wide variety of NLP techniques, ranging from models employing hand-engineered features \cite{shutova-etal-2010-metaphor,bulat-etal-2017-modelling}, to RNNs \cite{gao-etal-2018-neural,mao-etal-2019-end}, to more recently pre-trained language models \cite{choi_melbert_2021,flp-2022-figurative}, to mention just a few. Of particular interest for our work is the MelBERT model of \citet{choi_melbert_2021} whose design was founded on two linguistic theories of metaphor: the Metaphor Identification Procedure of \citet{Pragglejaz:2007}, instantiated in the model through contextualized vector representations of the target word and sentence, and the Selectional Preference Violation of \citet{Wilks:1975}, instantiated through a context independent representation of the target word.

In this paper, we present an architecture that, like MelBERT, is structured around the two modules, however the intended function of these modules is very different: one module is aimed at estimating the expectation (E) of a literal meaning given a context where the target word is missing, whereas the other module is aimed at estimating the realized (R) meaning of the actual word used in context. This ER architecture is motivated by the idea that the use of a metaphorical word leads to a violation of literal word expectations, i.e. {\it surprise}. Surprise has been postulated as a general core mechanism through which stories and music trigger emotion \cite{meyer:book61}. This notion of surprise has been recently shown to correlate with creative uses of language, such as humor and metaphor \cite{bunescu-uduehi-2022-distribution}.


\section{The Expectation-Realization Model}

\begin{figure*}
\centering
\includegraphics[width=1.0\textwidth]{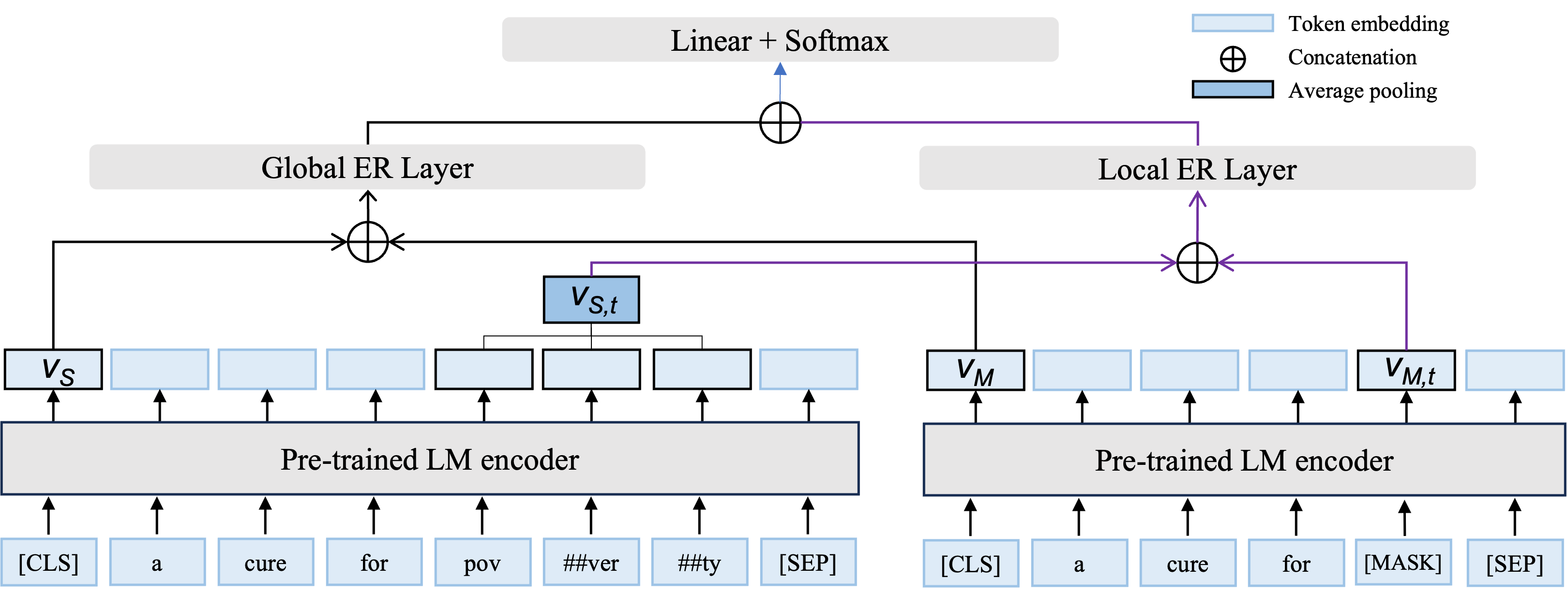}
\caption{Expectation-Realization Model architecture}
\label{er-model}
\end{figure*}

The architecture of the Expectation-Realization model for metaphor detection is shown in Figure~\ref{er-model}. To compute the realized (R) meaning $v_{S, t}$ of the actual word used in context, a copy of the Transformer encoder of a pre-trained language model (shown on the left) processes the input text $S$ where the target word at position $t$ is marked with a special token, akin to the encoding approach used in MelBERT. To compute the expectation (E) of a literal meaning $v_{M, t}$ given a context where the target word is missing, the same pre-trained language model (shown on the right) process the same input text $M$ where the target word is masked. Additionally, global expectation $v_{M}$ and realization $v_{S}$ representations are also computed at the sentence level using the embeddings for the special [CLS] token. The concatenation of the local target word ER embeddings and the sentence-level ER embeddings are passed through non-linear layers $f$ and $g$, respectively, to capture interactions between expectation and realization embeddings at word-level as $h_{local} = f[v_{M, t} ; v_{S, t}]$, and at sentence level as $h_{global} = g[v_{M} ; v_{S}]$. To enable a fair comparison with MelBERT, we instantiate the pre-trained Transformer encoder using RoBERTa base \cite{liu2019roberta}. The concatenated local and global ER representations are then used as input features to a logistic regression model that estimates the probability $\hat{y}$ that the target word is used metaphorically.
\begin{equation*}
\hat{y} = \sigma(w^\mathsf{T}[h_{local} ; h_{global}] + b) 
\end{equation*}
The ER model parameters together with the pre-trained LM parameters are trained and fine-tuned, respectively, in order to minimize a loss function $L^i = L^{i}_{CE} - L^{i}_{Sim}$ that contains a cross-entropy loss $L^{i}_{CE}$ and a similarity loss $L^{i}_{Sim}$ computed as:
\begin{eqnarray*}
L^{i}_{CE} & = & y_i \log \hat{y}_i + (1 - y_i) \log(1 - \hat{y}_i) \\
L^{i}_{Sim} & = & \alpha_1 \cos{(u_{M, t}, v_{M, t})} + \alpha_2 \cos{(u_{M}, v_{M})}
\end{eqnarray*}
where $y_i$ and $\hat{y}_i$ are the ground truth and predicted labels, respectively, for training sample $i$. The embeddings $u$ are obtained from the original pre-trained LM with fixed parameters, whereas the embeddigns $v$ are obtained from the fine-tuned LM. Consequently, the similarity loss encourages the fined-tuned LM to learn expectation embeddings $v$ that do not deviate much from the original embeddings produced by the pre-trained LM. The hyper-parameters $\alpha_1$ and $\alpha_2$ trade-off the global and local components of the similarity term within the overall loss. Given that most words in the vocabulary are used with their literal meaning most of the time, the similarity loss serves the purpose of anchoring the fine-tuned LM such that its expectation embeddings $v$ reflect a literal meaning of words.

\section{Experimental Evaluation}

We run evaluations on three English metaphor datasets: VUA20 \cite{chen-etal-2020-go}, TroFi \cite{birke-sarkar-2006-clustering} and LCC \cite{mohler-etal-2016-introducing}. Table \ref{dataset-stats} summarizes the statistics of the datasets used in our evaluations. The VU Amsterdam Metaphor Corpus (VUA) was released for metaphor detection shared tasks in 2018 and 2020. We use the VUA-20 for evaluating our models and baselines. The VUA-20 is an extension of VUA-18 and is split into training, validation and test datasets denoted by {VUA-20$_{tr}$} and {VUA-20$_{te}$} respectively. The examples in the VUA-20 dataset are sentences where selected words of the sentence are annotated as metaphorical or not. The LCC Metaphor dataset is a large, multilingual dataset of metaphor annotations created by a team of researchers at the  Language Computer Corporation (LCC). Each target word is annotated with a metaphoricity rating on a four-point scale [0, 3]. In our experiments we use a subset of the English dataset where examples with metaphoricity score of 3 are considered as positive and examples with metaphoricity score of 0 as negatives. The TroFi dataset consists of a collection of literal and nonliteral usage of 50 verbs which occur in 3,737 sentences selected from the Wall Street Journal (WSJ) corpus.

\begin{table}
\begin{center}
\begin{tabular}{c|cccc} \toprule
    {Dataset} & {\#words} & {\%M} & {\#Sent} & {Len} \\ \midrule
    {VUA-20$_{tr}$}  & 160,154 & 12.0 & 12,109 & 15.0 \\
    {VUA-20$_{te}$}  & 22,196 & 17.9 & 3,698 & 15.5 \\ \midrule
    {LCC}   & 5,646  & 28.9 & 5,390 & 28.9 \\ \midrule
    {TroFi}   & 3,737  & 43.5 & 3,737 & 28.3 \\ \bottomrule
    
\end{tabular}
\caption{\label{dataset-stats} Detailed statistics of datasets. \#words is the number of target words to be classified, \%M is the percentage of metaphorical words, \#Sent is the number of sentences, and Len is the average sentence length.}
\end{center}
\end{table}
 
We compare the performance of the ER model with two strong baseline models. First, we implement only the realization component of the ER model, using as input the sentence with the target word marked, as shown on the left of Figure~\ref{er-model}. We call this model R-SPV, as it is equivalent also with the SPV component of the MelBERT model. The second baseline is the MelBERT model itself.

\subsection{Generalization Settings}

The generalization performance of each of the 3 models is evaluated in three settings: {\it within distribution (WID)}, {\it out of distribution (OOD)}, and {\it novel} metaphor generalization. For the WID generalization, we randomly split the dataset into 10 folds and run 10-fold evaluation, where 9 folds are used for training and development, and 1 fold is used for testing, with the procedure repeated 10 times so that each folds gets to be used as a test fold. For OOD generalization, the 10 folds are created such that the lemmas of target words are disjoint across the folds. For the Novel generalization setting, we identify a subset of 237 positive examples within the LCC dataset that are novel or unconventional metaphors. The criteria for creating this subset were example with the highest metaphoricity score of 3.0 that were also rare according to a Google search, i.e. returning fewer that 25 search results. To complete the novel version of the dataset, negative examples are randomly sampled from the LCC dataset such that the ratio of positive to negatives for this novel dataset is similar to that of the original LCC dataset. For this evaluation we only compute the test performance on the novel subset of examples using  the models already trained on data from the within-distribution setting, ensuring that no novel test example has been used during training. Due to the imbalanced distribution of positive and negative examples in the datasets, we report only precision, recall and F1-score metrics. For 10-fold evaluation we report their micro-averages.

\subsection{Experimental details}

For the evaluations on VUA-20 dataset, we use the same hyperparameter settings from \cite{choi_melbert_2021} for training all models. 
For the LCC and TroFi experiments, the development dataset was used for determining the best hyperparameter settings. We use the same hyperparameter settings for all the models. The batch size and max sequence length were set at 32 and 150, respectively. We train for 12 epochs without dropout, and linearly increase the learning rate from 0 to 5e-5 in the first two epochs, after which we decreased it linearly to 0 during the remaining 10 epochs. The tuned similarity weights $\alpha_1$ and $\alpha_2$ were 1.0. for the within-distribution experiments and 0.0 for out-of-distribution experiments. Results are averaged over 5 runs with different random seeds.  The detailed ranges used for hyperparameters tuning are presented in Appendix \ref{appendix:hyper-tuning}.

\subsection{Empirical Results}

Tables \ref{vua-20-result}, \ref{lcc-result} and \ref{trofi-result} show the results of comparison of the ER Model against the baselines on the 3 datasets. Additionally, for the LCC and TroFi datasets in the WID setting we also report the performance of GPT-3.5 in a zero-shot setting, using the prompt detailed in Appendix~\ref{appendix:chatgpt-prompt}.

\begin{table}[t]
\begin{center}
\begin{tabular}{c|cccl} \toprule
    {Dataset} & {Model} & {Prec} & {Rec} & {F1} \\ \midrule
    \multirow{3}{*}{VUA-20} & {R-SPV} & 74.0 & 69.7 & 71.8 \\
    & {MelBERT} & 76.4 & 68.6 & 72.3 \\ 
    & {ER} & 75.3 & 70.2 & $72.6^{*\dag}$ \\ 
    \bottomrule
\end{tabular}
\caption{\label{vua-20-result} Performance comparison of ER model with baselines on VUA-20 dataset. * and \dag indicate significantly better F1 than R-SPV and MelBERT, respectively.}
\end{center}
\end{table}

\begin{table}
\begin{center}
\begin{tabular}{c|cccl} \toprule
    {Dataset} & {Model} & {Prec} & {Rec} & {F1} \\ \midrule
    \multirow{6}{2em}{LCC (WID)} & {R-SPV} & 86.2 & 83.9 & 85.0 \\
    & {MelBERT} & 86.1 & 83.8 & 84.9 \\ 
    & {GPT-3.5} & 67.9 & 64.6 & 66.2 \\
    & {ER} & 86.9 & 84.3 & $85.5^{*\dag}$ \\ \cline{2-5}
    & {ER-Ens (2)} & 87.5 & 85.0 & 86.2 \\
    & {ER-Ens (5)} & 87.7 & 85.3 & 86.5 \\
    \midrule
    \multirow{5}{2em}{LCC (OOD)} & {R-SPV} & 83.6 & 79.8 & 81.6 \\
    & {MelBERT} & 83.4 & 79.8 & 81.5 \\ 
    & {ER} & 84.0 & 80.6 & $82.2^{*\dag}$ \\ \cline{2-5}
    & {ER-Ens (2)} & 85.1 & 81.4 & 83.2 \\
    & {ER-Ens (5)} & 85.9 & 81.9 & 83.9 \\
    \midrule
    \multirow{5}{2em}{LCC (novel)} & {R-SPV} & 88.0 & 94.3 & 91.1 \\
    & {MelBERT} & 87.6 & 94.5 & 90.9 \\ 
    & {ER} & 88.8 & 95.1 & $91.8^{*\dag}$ \\ \cline{2-5}
    & {ER-Ens (2)} & 89.5 & 95.5 & 92.4 \\
    & {ER-Ens (5)} & 89.3 & 95.7 & 92.4 \\
    \bottomrule
\end{tabular}
\caption{\label{lcc-result} Performance comparison of ER model with baselines on LCC dataset. * and \dag indicate significantly better F1 than R-SPV and MelBERT, respectively.}
\end{center}
\end{table}

\begin{table}[t]
\begin{center}
\begin{tabular}{c|cccl} \toprule
    {Dataset} & {Model} & {Prec} & {Rec} & {F1} \\ \midrule
    \multirow{6}{2em}{TroFi (WID)} & {R-SPV} & 70.2 & 71.8 & 71.0 \\
    & {MelBERT} & 69.5 & 73.3 & 71.3 \\ 
    & {GPT-3.5} & 60.2 & 59.2 & 59.7 \\  
    & {ER} & 70.2 & 73.7 & $71.9^{*\dag}$ \\\cline{2-5}
    & {ER-Ens (2)} & 71.8 & 72.5 & 72.1 \\
    & {ER-Ens (5)} & 72.2 & 73.5 & 72.8 \\
    \midrule
    \multirow{5}{2em}{TroFi (OOD)} & {R-SPV} & 57.4 & 69.6 & 62.8 \\
    & {MelBERT} & 57.1 & 69.8 & 62.7 \\ 
    & {ER} & 57.0 & 70.5 & 63.0 \\ \cline{2-5}
    & {ER-Ens (2)} & 57.7 & 70.9 & 63.6 \\
    & {ER-Ens (5)} & 58.1 & 71.8 & 64.2 \\
    \bottomrule
\end{tabular}
\caption{\label{trofi-result} Performance comparison of ER model with baselines on TroFi dataset. * and \dag indicate significantly better F1 than R-SPV and MelBERT, respectively.}
\end{center}
\end{table}

For the the within distribution setting of VUA-20, LCC and TroFi, the ER model statistically significantly outperforms R-SPV and MelBERT, as determined through a one-tailed, paired t-test of significance at $p < 0.05$ level. In the Novel evaluation setting for the LCC dataset, the ER model again outperforms both baselines. These results show the ER model is generally better at identifying metaphorical uses of words, especially for novel metaphors. The VUA-20 results are notably lower than the LLC results for all methods. Error analysis revealed that almost any non-literal use of a word is annotated as a positive example in VUA, including idioms. Therefore, the patterns are more complicated. Idioms, in particular, lack any clear pattern, hence they require memorization, which may explain the much lower OOD performance.

For the out-of-distribution evaluation on LCC and TroFi dataset, the ER model on average performs better than both R-SPV and MelBERT, with the comparison on LCC being statistically significant. The results from the OOD settings however, show a significant drop compared to the within distribution setup with the result being less worse for LCC than TroFi because of the more diverse nature of the target words in the LCC dataset. This drop in performance in the OOD scenario suggests that the models rely on some form of memorization, which is detrimental to identifying metaphors that use unseen words. The nature of the TroFi dataset makes the OOD generalization even worse, as the dataset contains only 50 words and thus the model has limited diversity in terms of target metaphorical words. Overall, the ER model shows its competitiveness at better identifying unseen words that are used metaphorically at test time.

For zero-shot evaluation of GPT-3.5, a logistic regression classifier is trained using GPT-3.5's responses to questions 2 to 14 of the prompt as features. The results reveal that GPT-3.5 exhibits the lowest performance on both LCC and TroFI under the WID settings amongst the models. This indicates that GPT-3.5 struggles to accurately identify metaphors and handle the lexical complexity associated with metaphorical language. Future research could improve this by fine-tuning the model on metaphor-focused datasets and enhancing its commonsense reasoning capabilities.

Lastly, ensembles ER-Ens of 2 or 5 ER models are shown to further improve the metaphor detection performance. 

\section{Conclusion}

We introduced a new model for metaphor detection that is rooted in the hypothesis that the non-literal use of a word triggers surprise, or violation of expectations of literal word usage in a given context. The proposed architecture therefore contains a component that aims to compute expectations of literal word usage induced by the context, and a component that aims to compute realization representations of actual word usage in the context. Through extensive experiments on 3 metaphor datasets in 3 different generalization settings, the ER model is shown to be competitive and to often outperform the state-of-the-art MelBERT as well as another strong baseline model. 

\bibliography{emnlp2023}
\bibliographystyle{acl_natbib}



\appendix
\section{Hyperparameter Tuning} \label{appendix:hyper-tuning}

Details for the hyperparameter tuning for the models and dataset are presented in Table \ref{table:hyp-tuning}.

\begin{table}[h]
\begin{center}
\begin{tabular}{l|l} \toprule
    {Hyperparameter} & {Tuning values} \\ \midrule
    {learning rate}  & [1e-5, 2e-5, 3e-5, 4e-5, 5e-5] \\
    {dropout ratio}  & [0.0, 0.10, 0.20, 0.25, 0.40, 0.50] \\
    {similarity weight $\alpha$}  & [0, 0.5, 1, 2, 4] \\
    {hidden dims}   & [[768], [768,768], [768,768,1]] \\
    {hidden activation}   & [None, relu] \\
    {optimizer}  & [Adam] \\
    {train batch size}  & [32] \\
    
     \bottomrule
\end{tabular}
\caption{\label{hyper-tuning} Hyperparameters tuning range used in experiments. For the similarity weight, $\alpha = \alpha_1 = \alpha_2.$}
\label{table:hyp-tuning}
\end{center}
\end{table}

\section{GPT-3.5 prompt template} \label{appendix:chatgpt-prompt}

The sample prompt we used to query GPT-3.5 is shown below: \\

\noindent You are a professional linguist. For the text below, answer precisely the following questions. Only print out a Python list containing your answers. \\

\noindent text: The sun *walked* between the clouds. \\

\noindent 1. What word is emphasized? \\
2. Is the emphasized word "walked" used literally in the text? Yes or No? \\
3. Is the emphasized word "walked" used figuratively in the text? Yes or No? \\
4. Is the emphasized word "walked" used metaphorically in this text? Yes or No? \\
5. Is the emphasized word "walked" used with its literal meaning in the text? Yes or No? \\
6. Is the emphasized word "walked" used with its most common literal meaning in this text? Yes or No? \\
7. Is the emphasized word "walked" used with a concrete meaning in the text? Yes or No? \\
8. Is the emphasized word "walked" used with a physical meaning in the text? Yes or No? \\
9. Is the emphasized word "walked" used with its conventional meaning in the text? Yes or No? \\
10. Is the emphasized word "walked" used with its most common meaning in this text? Yes or No? \\
11. Is the emphasized word "walked" used with its original (oldest) meaning in this text? Yes or No? \\
12. Is the emphasized word "walked" part of a metaphorical expression in the text? Yes or No? \\
13. Is the emphasized word "walked" part of an idiomatic expression in the text? Yes or No? \\
14. Is the emphasized word "walked" part of a multiword expression in the text? Yes or No? \\

\end{document}